\documentclass[conference]{IEEEtran}
\IEEEoverridecommandlockouts

\usepackage{cite}
\usepackage{amsmath,amssymb,amsfonts}
\usepackage{algorithmic}
\usepackage{graphicx} 
\usepackage{textcomp}
\usepackage{xcolor}   
\definecolor{LightGray}{gray}{0.9}
\usepackage{booktabs}
\usepackage{multirow}
\usepackage{adjustbox}
\usepackage{listings}
\usepackage{tikz}     

\definecolor{bggray}{rgb}{0.95,0.95,0.95}

\usepackage[most]{tcolorbox}

\usepackage{fancyvrb}
\usepackage{minted}

\usepackage{fancyhdr} 

\usepackage[colorlinks]{hyperref}

\def\BibTeX{{\rm B\kern-.05em{\sc i\kern-.025em b}\kern-.08em
    T\kern-.1667em\lower.7ex\hbox{E}\kern-.125emX}}

\begin{document}

\title{Turk-LettuceDetect: A Hallucination Detection Models for Turkish RAG Applications}

\author{\IEEEauthorblockN{\textit{Authors Hidden for Review}}}
\author{
    \IEEEauthorblockN{
        Selva Taş, Mahmut El Huseyni, Özay Ezerceli, Reyhan Bayraktar, Fatma Betül Terzioğlu
    }
    \IEEEauthorblockA{
        Newmind AI\\
        Istanbul, Türkiye\\
        \texttt{\{stas,mehussieni,oezerceli,rbayraktar,fbterzioglu@newmind.ai\}@newmind.ai}
    }
}
\maketitle


\begin{abstract}
The widespread adoption of Large Language Models (LLMs) has been hindered by their tendency to hallucinate, generating plausible but factually incorrect information. While Retrieval-Augmented Generation (RAG) systems attempt to address this issue by grounding responses in external knowledge, hallucination remains a persistent challenge, particularly for morphologically complex, low-resource languages like Turkish. This paper introduces Turk-LettuceDetect, the first suite of hallucination detection models specifically designed for Turkish RAG applications. Building on the LettuceDetect framework, we formulate hallucination detection as a token-level classification task and fine-tune three distinct encoder architectures: a Turkish-specific ModernBERT, TurkEmbed4STS, and multilingual EuroBERT. These models were trained on a machine-translated version of the RAGTruth benchmark dataset containing 17,790 instances across question answering, data-to-text generation, and summarization tasks. Our experimental results show that the ModernBERT-based model achieves an F1-score of 0.7266 on the complete test set, with particularly strong performance on structured tasks. The models maintain computational efficiency while supporting long contexts up to 8,192 tokens, making them suitable for real-time deployment. Comparative analysis reveals that while state-of-the-art LLMs demonstrate high recall, they suffer from low precision due to over-generation of hallucinated content, underscoring the necessity of specialized detection mechanisms. By releasing our models and translated dataset, this work addresses a critical gap in multilingual NLP and establishes a foundation for developing more reliable and trustworthy AI applications for Turkish and other  languages.

\end{abstract}


\section{Introduction}

The rapid advancement of Large Language Models (LLMs) has revolutionized natural language processing (NLP) applications, demonstrating remarkable capabilities in text generation, question answering, and reasoning tasks \cite{Qin2024}. However, their propensity to generate plausible-sounding but factually incorrect information, commonly referred to as hallucinations, remains a critical challenge that limits their deployment in real-world applications where accuracy and reliability are paramount \cite{Zhang2023,Ji2023}.

Retrieval-Augmented Generation (RAG) has emerged as a promising paradigm to mitigate hallucination issues by grounding LLM responses in external knowledge sources \cite{Lewis2020,Gao2024}. By retrieving relevant documents and incorporating them as context during generation, RAG systems aim to anchor model outputs in factual information rather than relying solely on parametric knowledge. Despite these architectural improvements, empirical studies have shown that RAG systems still exhibit significant hallucination rates, particularly when dealing with complex queries or when the retrieved context is incomplete or contradictory \cite{Niu2024,Shuster2021}.

The challenge of hallucination detection becomes even more pronounced in low-resource languages where training data is scarce and evaluation benchmarks are limited. Turkish, despite being spoken by over 80 million people worldwide, represents one such language where robust hallucination detection approachs are critically needed but underexplored \cite{Benkirane2024}. The linguistic complexity of Turkish, characterized by its agglutinative morphology and rich inflectional system, poses additional challenges for accurate hallucination detection compared to morphologically simpler languages like English \cite{Oflazer2003}.

Existing hallucination detection approaches can be broadly categorized into two paradigms: prompt-based methods that leverage large language models as judges \cite{Min2023,Chern2023}, and fine-tuned encoder-based models that perform token-level classification \cite{Belyi2025,Azaria2023}. While prompt-based methods offer flexibility and can achieve high accuracy, they suffer from computational inefficiency and high inference costs due to their reliance on large models. Conversely, traditional encoder-based approaches, while computationally efficient, are constrained by limited context windows that prevent them from processing long documents typical in RAG applications.

Recent advances in encoder architectures have begun to address these limitations. ModernBERT \cite{Warner2024}, building upon the foundational BERT architecture \cite{Devlin2019}, incorporates several key innovations including rotary positional embeddings (RoPE) \cite{Su2024} and local-global attention mechanisms that enable processing of sequences up to 8,192 tokens. These architectural improvements make ModernBERT particularly well-suited for hallucination detection in RAG systems, where long context understanding is essential for accurate verification of generated content against source documents.

Drawing from the recently developed LettuceDetect framework \cite{Kovacs2025}, we introduce Turk-LettuceDetect, a tailored variant optimized for Turkish RAG systems. The original LettuceDetect framework showed substantial advancements in identifying hallucinations within English content and was subsequently expanded to accommodate various languages such as German, French, Spanish, Italian, Polish, and Chinese through EuroBERT integration. It achieved a 79.22\% F1 score on the RAGTruth benchmark, which marks a 14.8\% enhancement compared to earlier encoder-based methods in English. Nevertheless, a significant void persists in cross-linguistic hallucination detection, especially for languages with complex morphology and limited resources such as Turkish.

Our approach extends the core LettuceDetect methodology by adapting three different model base such as ModernBERT, gte-multilingual based TurkEmbed4STS and EuroBERT on token classification to handle the unique linguistic characteristics of Turkish while maintaining the computational efficiency of the original framework. We fine-tuned Turkish-specific models on machine translated versions of the RAGTruth dataset \cite{Niu2024}, formulating the task as a binary token classification problem where each token in the generated answer is labeled as either supported or hallucinated based on the given context and question.

The main contributions of this paper are threefold:
\begin{enumerate}
    \item We present \textit{Turk-LettuceDetect}, the first adaptation of the LettuceDetect for Turkish RAG applications with three open-source models, addressing a significant gap in multilingual hallucination detection research.
    
    \item We demonstrate that the ModernBERT-based architecture can maintain competitive performance when adapted to Turkish text, preserving the computational efficiency advantages of the original framework while handling the morphological complexity of Turkish.
    
    \item We provide extensive experimental validation across multiple task types including question answering, data-to-text generation, and summarization, showing that our Turkish-specific models achieve consistent performance improvements over baseline multilingual approaches.
    
    \item The translated Turkish-RAGTruth dataset and the fine-tuned hallucination-detection models have been released under an open-source license to support and accelerate future research.
    
\end{enumerate}

The multilingual extensions built on diverse model backbones demonstrate the approach’s adaptability to cross-lingual contexts, thereby paving the way for broader multilingual deployment of the LettuceDetect methodology.

The remainder of this paper is organized as follows: Section~\ref{sec:related_work} reviews related work; Section~\ref{sec:methodology} describes the dataset and methodology; Section~\ref{sec:results} presents experimental results and comparisons; Section~\ref{sec:conclusion} concludes the paper with a discussion of future directions.

\section{Related Work}
\label{sec:related_work}
Hallucinations are prevalent in both general-purpose LLMs and RAG systems \cite{Zhang2023,Ji2023}. These phenomena have been extensively studied in recent literature, including the comprehensive survey by Ahadian and Guan \cite{Ahadian2024}, which identifies two major causes: (1) training data biases and memorization effects, and (2) decoding time errors such as sampling artifacts or insufficient grounding.

In RAG systems, hallucinations can arise due to misinterpretation of retrieved evidence, retrieval inaccuracies, or misalignment between retrieved documents and the user’s query \cite{Niu2024,Gao2024}. Several mitigation strategies have been proposed, broadly categorized into \textit{prompt-based} and \textit{model-based} approaches.

Prompt-based methods leverage LLMs as judges to evaluate factual consistency using structured prompts or reasoning chains \cite{Min2023,Chern2023}. While flexible and easy to implement, they suffer from high computational cost, lack of interpretability, and inconsistent judgments across model versions \cite{Azaria2023}.

Model-based approaches, on the other hand, fine-tune smaller encoder-based models to detect hallucinated spans directly. These methods offer better efficiency and reproducibility, making them suitable for real-time deployment. Notable examples include \textit{Luna} \cite{Belyi2025}, which uses a classification head over contextual embeddings to identify unsupported content, and \textit{LettuceDetect} \cite{Kovacs2025}, which extends this idea with token-level classification and achieves significant improvements in precision and recall over prior encoder-based baselines.

ModernBERT \cite{Warner2024}, building upon BERT's foundational design \cite{Devlin2019}, incorporates several innovations such as rotary positional embeddings (RoPE) \cite{Su2024} and local-global attention mechanisms, enabling it to process sequences up to 8,192 tokens. This makes it especially well-suited for hallucination detection in RAG systems, where long context understanding is crucial for accurate verification against source documents.

The development of robust and innovative Turkish embedding models for downstream tasks such as semantic textual similarity (STS) and information retrieval necessitates the exploration and deployment of novel backbone architectures tailored to the unique characteristics of the Turkish language. Recent contributions, including TurkEmbed4STS and TurkEmbed4Retrieval \cite{Ezerceli2025}, have demonstrated the critical importance of domain-specific fine-tuning for Turkish retrieval applications, while simultaneously highlighting the need for architectural innovations to achieve optimal performance across diverse downstream tasks. Building upon these foundational insights, our work extends this paradigm by adapting state-of-the-art encoder architectures specifically for hallucination detection in Turkish RAG applications, thereby addressing both the inherent morphological complexity of Turkish and the limitations associated with multilingual transfer learning approaches.

Unlike traditional NLI-based approaches, our method is entirely trained on the machine-translated version of RAGTruth dataset \cite{Niu2024}, using ModernBERT, GTE-multilingual-base and EuroBERT base architectures. We also incorporate multilingual extensions of the dataset, translating it using Gemma3-27b-it \cite{Team2024} for cross-lingual generalization.

Our results show that efficient encoder-based hallucination detection can rival prompt-based methods while maintaining inference speed and scalability, an important step toward deploying RAG systems in low-resource, high-stakes environments.

\section{Methodology}\label{sec:methodology}

This section presents our methodology for hallucination detection within RAG systems. We employed the LettuceDetect framework to train hallucination detection models using three distinct Turkish-supported encoder architectures. The first model, ModernBERT-base-tr, was specifically fine-tuned for this research on Turkish Natural Language Inference (NLI) and STS downstream tasks, following the established paradigms and methodologies demonstrated by the TurkEmbed approach. The second model is TurkEmbed4STS model (gte-multilingual-based base) and the third model utilized was euroBERT, an established pre-existing multilingual encoder. This experimental design enabled a comprehensive evaluation of hallucination detection capabilities across diverse encoder backbones within the Turkish linguistic context.

\subsection{Dataset}

We utilized the RAGTruth dataset \cite{Niu2024} for both training and evaluation of our models. RAGTruth constitutes the first large-scale benchmark specifically constructed to evaluate hallucination phenomena in RAG settings. The dataset comprises annotated 17.790 train, 2.700 test instances, covering three distinct tasks: question answering, data-to-text generation, and summarization.

For the question answering task, instances were sourced from the MS MARCO dataset \cite{Bajaj2018}. Each question was matched with up to three context passages retrieved via information retrieval techniques, and LLMs were prompted to generate answers based on these contexts. In the data-to-text generation task, models produced reviews for businesses selected from the Yelp Open Dataset \cite{Yelp2021}. For the news summarization task, documents were randomly sampled from the training set of the CNN/Daily Mail corpus \cite{See2017}, and the LLMs were asked to generate abstractive summaries accordingly.

A diverse set of LLMs was employed for response generation, including GPT-4-0613 \cite{OpenAI2024}, Mistral-7B-Instruct \cite{Jiang2023}, and several models from the LLaMA family, such as LLaMA2-7B-Chat and LLaMA2-13B-Chat \cite{Grattafiori2024}. Each data point in the dataset includes responses from six different models, allowing for multi-model comparison at the example level. All samples were manually annotated by human experts, who identified hallucinated spans within the model outputs and provided rationale for their decisions. The dataset further classifies hallucinations into four distinct categories: Evident Conflict, Subtle Conflict, Evident Introduction of Baseless Information, and Subtle Introduction of Baseless Information. However, for the purposes of our model training, we simplified this classification to a binary hallucination detection task, disregarding the specific hallucination types.

An analysis of token lengths within the dataset revealed a mean input length of 801 tokens, a median of 741, with lengths ranging from 194 to 2,632 tokens. These findings underscore the need for long-context language models which has context window bigger than 4096 tokens to effectively identify hallucinations, especially in lengthy and context-rich inputs.

\subsection{Multilingual Extension of RAGTruth Dataset}

To address the monolingual limitation of the RAGTruth dataset, we developed a multilingual extension by translating its content into multiple target languages by following the way thet is translated to other european languages. This subsection details the translation protocols, and a sample of the translated dataset.

\subsubsection{Translation Methodology}

The translation pipeline utilized the \texttt{google/gemma-3-27b-it} model, executed via vLLM on a single NVIDIA A100 GPU. This configuration supported parallel processing of approximately 30 examples, with a full translation pass for target language completed in roughly 12 hours.

\subsubsection{Translation Protocols} 

The translation protocols in this study were designed to maintain structural integrity while ensuring accurate cross-lingual transfer of both content and metadata. Two distinct translation procedures were implemented: one for answer content and another for prompt instructions, each tailored to handle specific linguistic and structural requirements.

\begin{tcolorbox}[ 
    colback=bggray,
    boxrule=0.5pt,
    arc=0pt,
    outer arc=0pt,
    width=\columnwidth, 
    fonttitle=\bfseries,
    title=Core Translation Prompt
]
\begin{minted}[
    breaklines=true,
    fontsize=\footnotesize
]{text}
Translate the following text from 
{source_lang} to {target_lang}. If the 
original text contains <HAL> tags, translate 
the content inside <HAL> tags and ensure the 
number of the <HAL> tags remain exactly the 
same in the output. If the original text does 
not contain <HAL> tags, just translate the 
text. Do NOT add any <HAL> tags if they were 
not in the original text. Do NOT remove any 
<HAL> tags that were in the original text. Do 
not include any additional sentences 
summarizing or explaining the translation. 
Your output should be just the translated 
text, nothing else.
\end{minted}
\end{tcolorbox}


\paragraph{Answer Translation Protocol}
The answer translation protocol was specifically designed to handle hallucination-annotated content while preserving the evaluation framework's structural requirements. When translating answers from source language to target language, the following systematic approach was employed:

\begin{itemize}
    \item \textbf{Tag Preservation:} The exact number and positioning of \texttt{<HAL>} tags from the source text were maintained, with translation applied exclusively to the content within these tags. This ensures that hallucination annotations remain consistent across languages for comparative analysis.
    \item \textbf{Content-Only Translation:} For text segments without \texttt{<HAL>} tags, direct translation was performed while maintaining the original semantic meaning and contextual nuances.
    \item \textbf{Structural Integrity:} No \texttt{<HAL>} tags were introduced unless present in the source material, and no existing tags were removed, ensuring the hallucination detection framework remains intact across all translated versions.
    \item \textbf{Output Specification:} Only the translated text was produced, excluding any meta-commentary, explanations, or translation notes that could interfere with subsequent automated processing.
\end{itemize}

\paragraph{Prompt Translation Protocol}
The prompt translation protocol addressed the challenge of translating instruction sets while maintaining their functional effectiveness across different linguistic contexts. For translating prompts from source language to target language, the following specialized methodology was implemented:

\begin{itemize}
    \item \textbf{Comprehensive Content Translation:} All prompt components were translated, including both natural language instructions and structured elements such as JSON objects, where both keys and values underwent linguistic transformation to ensure cultural and linguistic appropriateness.
    \item \textbf{Functional Equivalence:} Translation prioritized maintaining the prompt's intended function and directive clarity rather than literal word-for-word conversion, ensuring that the translated prompts elicit equivalent responses from language models.
    \item \textbf{Clean Output Generation:} The translation process produced only the target language prompt without supplementary explanatory text, facilitating direct integration into the experimental pipeline.
\end{itemize}

These protocols collectively ensure that the cross-lingual evaluation maintains both the semantic integrity of the original content and the structural requirements necessary for consistent hallucination detection across multiple language contexts.

\subsubsection{Sample Translated Data}

Table~\ref{tab:ragtruth} provides an example of translated RAGTruth data, including a question, reference passage, response, and associated annotation, demonstrating the application of the translation protocols.

\begin{table}[ht]
\centering
\caption{Example of Translated RAGTruth Data}
\label{tab:ragtruth}
\scriptsize
\resizebox{\columnwidth}{!}{%
\begin{tabular}{@{}p{2.2cm} p{5.8cm}@{}}
\toprule
\textbf{PROMPT} & \textbf{Summarize the following news within 116 words:} \\
& Seventy years ago, Anne Frank died of typhus in a Nazi concentration camp at the age of 15. Just two weeks after her supposed death on March 31, 1945, the Bergen-Belsen concentration camp where she had been imprisoned was liberated -- timing that showed how close the Jewish diarist had been to surviving the Holocaust. But new research released by the Anne Frank House shows that Anne and her older sister, Margot Frank, died at least a month earlier than previously thought. \dots \\[3pt]
\textbf{ANSWER} & New research conducted by the Anne Frank House has revealed that Anne Frank and her sister Margot likely died in the Bergen-Belsen concentration camp at least a month earlier than previously believed. The researchers examined archives of the Red Cross, the International Training Service, and the Bergen-Belsen Memorial, as well as testimonies from survivors. \dots \\[3pt]
\textbf{ANNOTATION} &
\textbf{labels:}\{start: 636, end: 653, text: ``February 7, 2022.'', type: Evident Conflict\}, \{start: 871, end: 969, text: ``has prompted...'', type: Evident Baseless Info\}, \{start: 607, end: 646, text: ``believed to have died before February 7'', type: Evident Conflict\} \\
& \textbf{Type:} Summary \\
& \textbf{split:} train \\
\bottomrule
\end{tabular}%
}

\vspace{1em} 

\resizebox{\columnwidth}{!}{%
\begin{tabular}{@{}p{2.2cm} p{5.8cm}@{}}
\toprule
\textbf{PROMPT} & \textbf{Aşağıdaki haberi 116 kelimeyle özetleyin:} \\
& Yetmiş yıl önce Anne Frank, Nazi toplama kampında 15 yaşında tifo nedeniyle öldü. 31 Mart 1945'te öldüğü varsayılan tarihten sadece iki hafta sonra, tutuklu olduğu Bergen-Belsen toplama kampı kurtarıldı; bu zamanlama, Yahudi günlük yazarının Holokost'tan kurtulmaya ne kadar yakın olduğunu gösterdi. Ancak Anne Frank Evi tarafından yayınlanan yeni araştırmalar, Anne ve ablası Margot Frank'ın daha önce düşünüldüğünden en az bir ay önce öldüğünü gösteriyor. \dots \\[3pt]
\textbf{ANSWER} & Anne Frank Evi tarafından yapılan yeni bir araştırma, Anne Frank ve kız kardeşi Margot'un daha önce inanıldığından en az bir ay önce Bergen-Belsen toplama kampında ölmüş olabileceğini ortaya koydu. Araştırmacılar, Kızıl Haç, Uluslararası Eğitim Hizmeti ve Bergen-Belsen Anıtı arşivlerinin yanı sıra kurtulanların ifadelerini inceledi. \dots \\[3pt]
\textbf{ANNOTATION} &
\textbf{labels:}\{``start'': 545, ``end'': 596, ``label'': ``Evident Conflict''\}, \{``start'': 824, ``end'': 906, ``label'': ``Evident Baseless Info''\}\\
& \textbf{Type:} Summary \\
& \textbf{split:} train \\
& \textbf{language:} tr \\
\bottomrule
\end{tabular}%
}
\end{table}

\subsection{Model Architecture}
We propose a token-level hallucination detection pipeline based on three transformer-based encoder architectures: ModernBERT \cite{Warner2024}, TurkEmbed4STS, and euroBERT. The framework formulates hallucination detection as a binary token classification task, where each token in the generated response is classified as either supported or unsupported by the provided context.

The classification head produces binary predictions for each token in the input sequence. Training minimizes the cross-entropy loss across all tokens in the sequence, with appropriate masking for special tokens and padding.

Unlike previous encoder-based approaches that require NLI pretraining or auxiliary task transfers, our architecture operates directly on the base transformer representations. This design choice eliminates the need for multi-stage training pipelines and cross-task knowledge transfer, resulting in a more streamlined and reproducible system. The architecture's simplicity facilitates deployment in production environments while maintaining competitive performance on hallucination detection tasks. The token-level granularity enables fine-grained identification of hallucinated content, providing interpretable outputs for downstream applications requiring precise localization of factual inconsistencies.

\subsection{Training Configuration}

We fine-tuned three transformer-based models ModernBERT-base-tr, TurkEmbed4STS, and lettucedect-210m-eurobert-tr-v1 as token-level classifiers on the RAGTruth dataset. To focus on answer tokens, context and question tokens were masked (label = -100), while answer tokens were assigned labels of 0 (supported) or 1 (hallucinated) based on human annotations. Each model underwent independent fine-tuning with identical supervision to enable a fair comparison of their token-level hallucination detection performance under consistent data and labeling conditions.

Training was conducted over 6 epochs with a learning rate of $1 \times e^{-5}$ and a batch size 4. All training was performed on an NVIDIA A100 40GB GPU. Each epoch took around 20 minutes and whole training took 2 hours for each model.  

\section{Evaluation}\label{sec:results}

This section presents a comprehensive evaluation of our hallucination detection models across multiple tasks and settings. We assess model performance on the RAGTruth dataset test split and analyze the hallucination behavior of various decoder-only models and encoders to demonstrate the effectiveness of our approach.

\subsection{Experimental Setup}

We conducted a comprehensive evaluation of our models using the test split of the RAGTruth dataset, encompassing three core task types: question answering (QA), data-to-text generation, and summarization. To ensure a nuanced assessment, model performance was analyzed at both the example level and the token level, which provides a fine-grained perspective on model behavior within each generated text.

\paragraph{Evaluation Metrics} To rigorously assess hallucination detection capabilities, we employed a multi-metric evaluation framework:
\begin{itemize}
    \item Precision: The proportion of predicted hallucinated (or supported) instances that are actually correct, reflecting the model’s ability to avoid false positives.
    \item Recall: The proportion of actual hallucinated (or supported) instances that are correctly identified by the model, capturing sensitivity to true cases.
    \item Macro F1-Score: Calculated separately for hallucinated and supported instances, this metric provides a balanced measure of precision and recall, ensuring fair assessment across both prediction types.
    \item Area Under the Receiver Operating Characteristic Curve (AUROC): Summarizes model performance across all classification thresholds, offering an aggregate measure of discriminative ability.
\end{itemize}

This multi-level, multi-metric approach enables a robust and multifaceted evaluation of model performance across diverse aspects of hallucination detection in natural language generation tasks. All experiments were conducted using the latest stable versions of the lettucedetect and datasets libraries, and computations were performed on an NVIDIA A100 GPU with 40GB of memory, ensuring both methodological transparency and reproducibility.

\subsection{Comparative Model Performance and LLM Hallucination Behavior}
We evaluated three encoder-based models for hallucination detection in Turkish RAGTruth:
\begin{itemize}

\item \textbf{modernbert-base-tr-uncased-stsb-HD}: A Turkish-specific ModernBERT variant fine-tuned on hallucination detection.
    
\item  \textbf{TurkEmbed4STS-HallucinationDetection}: A Turkish embedding model optimized for semantic similarity and adapted to hallucination detection.
    
\item  \textbf{lettucedect-210m-eurobert-tr-v1}: A multilingual EuroBERT variant fine-tuned on hallucination detection.
\end{itemize}

\begin{figure*}[t]
\centering
\includegraphics[width=0.95\linewidth]{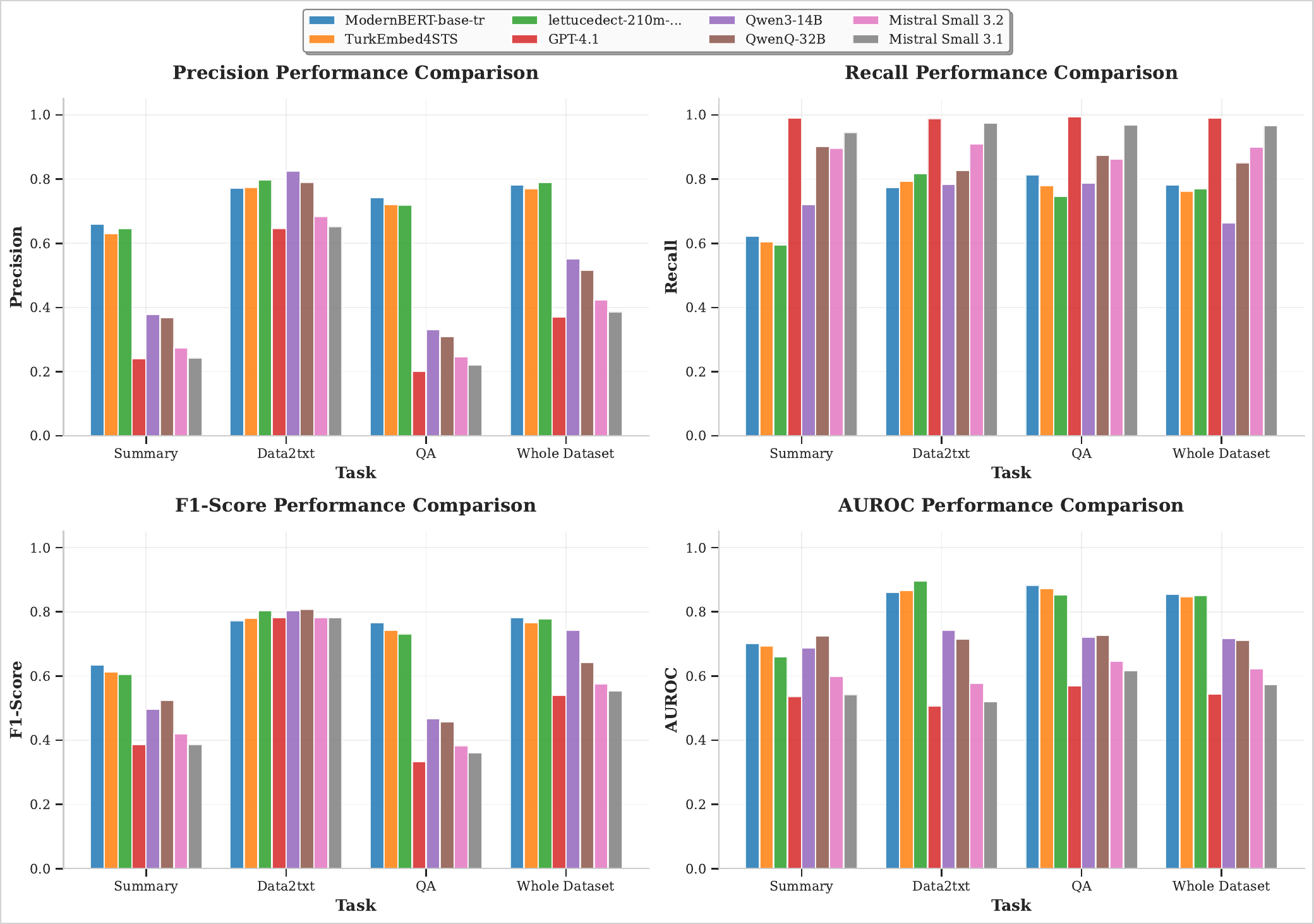}
\caption{Performance of example-level hallucination detection across models}
\label{fig:example_level}
\end{figure*}

The results, summarized in Figure~\ref{fig:example_level}, reveal distinct performance patterns across models and tasks. Among encoder-based models, \textit{TurkEmbed4STS-HallucinationDetection} shows the most consistent behavior, with relatively balanced precision and recall across all tasks. \textit{LettuceDetect-210m-EuroBERT-Tr-v1} achieves the highest AUROC in data-to-text generation (0.8966), indicating strong discriminative power between supported and hallucinated content. Meanwhile, \textit{modernbert-base-tr-uncased-stsb-HD} performs best in QA (AUROC = 0.8833), suggesting that Turkish-specific pretraining improves performance in structured tasks.

However, all encoder-based models show reduced effectiveness in summarization, particularly in detecting hallucinated tokens, with F1 scores below 0.65. This highlights the need for improved handling of abstractive generation in future hallucination detection frameworks.

When analyzing LLM hallucination behavior, we observe that GPT-4.1 and Mistral models achieve high recall (up to 0.9938), indicating a strong tendency to generate content flagged as hallucinated. However, their precision remains low, suggesting over-generation or systematic hallucination patterns.

In contrast, Qwen3-14B demonstrates the best overall balance, achieving the highest F1-score (0.7429) on the full dataset. It performs particularly well in data-to-text generation (precision = 0.8255), while QwenQ-32B excels in QA with the highest AUROC (0.7267).

These findings underscore the critical necessity of implementing specialized hallucination detection frameworks, such as Lettuce Detect, in conjunction with fine-tuned language models particularly within production-level Turkish RAG applications where maintaining factual accuracy and computational efficiency represents paramount operational requirements.

\begin{table}[t]
\centering
\caption{Performance of Token-Level Hallucination Detection Across Models}
\label{tab:Token-Level}
\begin{tabular}{lcccccc}
\toprule
\textbf{Model} & \textbf{Task Type} & \textbf{Precision} & \textbf{Recall} & \textbf{F1-Score} & \textbf{AUROC} \\
\midrule
\multicolumn{6}{l}{\textbf{ModernBERT-base-tr}} \\
& Summary & \textbf{0.6935} & \textbf{0.5705} & \textbf{0.6007} & \textbf{0.5705} \\
& Data2txt & 0.7652 & 0.7182 & 0.7391 & 0.7182 \\
& QA & \textbf{0.7642} & \textbf{0.7536} & \textbf{0.7588} & \textbf{0.7536} \\
& Whole Dataset & \textbf{0.7583} & \textbf{0.7024} & \textbf{0.7266} & \textbf{0.7024} \\
\midrule
\multicolumn{6}{l}{\textbf{TurkEmbed4STS}} \\
& Summary & 0.6325 & 0.5656 & 0.5862 & 0.5656 \\
& Data2txt & 0.7397 & \textbf{0.7333} & 0.7365 & \textbf{0.7333} \\
& QA & 0.7378 & 0.7382 & 0.7380 & 0.7382 \\
& Whole Dataset & 0.7268 & 0.7014 & 0.7132 & 0.7014 \\
\midrule
\multicolumn{6}{l}{\textbf{lettucedect-210m-eurobert-tr}} \\
& Summary & 0.6465 & 0.5546 & 0.5771 & 0.5546 \\
& Data2txt & \textbf{0.7866} & 0.7218 & \textbf{0.7496} & 0.7218 \\
& QA & 0.7388 & 0.7262 & 0.7323 & 0.7262 \\
& Whole Dataset & 0.7511 & 0.6908 & 0.7163 & 0.6908 \\
\bottomrule
\end{tabular}
\end{table}

\subsection{Token-Level Hallucination Detection}
\label{subsec:token_level_insights}

Table~\ref{tab:Token-Level} presents a detailed breakdown of hallucination detection performance at the token level across different models and task types. The results reveal several important patterns in how well models distinguish between supported and hallucinated content within generated responses.

First, all models demonstrate relatively strong performance on supported tokens (Class 0), with precision values above 0.63 and recall above 0.72. This indicates that the models are generally effective at identifying content that is grounded in the provided context. However, there is a noticeable drop in performance when detecting hallucinated tokens (Class 1), particularly in summarization tasks. For example, \textbf{modernbert-base-tr-uncased-stsb-HD} achieves only 0.6935 precision in summarization, suggesting that hallucinated content is more challenging to identify in this domain.

Second, among the evaluated models, \textbf{modernbert-base-tr-uncased-stsb-HD} consistently delivers the most balanced performance across both classes. In question answering, it achieves a precision of 0.7642 and a recall of 0.7536, showing its ability to detect hallucinations without sacrificing support accuracy. This supports our earlier findings that multilingual pre-training and fine-tuning enhances robustness and generalization across RAG tasks.

Third, while \textbf{modernbert-base-tr-uncased-stsb-HD} performs best in summary, QA and whole dataset it shows weaker hallucination detection in data2txt, where best precision and F1-score achieved by \textbf{lettucedect-210m-eurobert-tr} and best recall and AUROC achieved by \textbf{TurkEmbed4STS} models. This suggests that Turkish-specific pretraining helps in structured tasks but may require further adaptation for abstractive generation settings.

Interestingly, \textbf{TurkEmbed4STS} exhibits the most consistent behavior across all tasks, with precision and recall values staying relatively close. Although not always top-performing, it avoids extreme imbalances between supported and hallucinated class detection, making it a promising candidate for applications requiring stability over peak performance.

Finally, the overall token-level scores demonstrate that encoder-based models can reliably detect hallucinations in RAG outputs. More importantly, the high AUROC values confirm that these models possess strong discriminative power between supported and hallucinated content, even in the presence of class imbalance.

\section{Conclusion}
\label{sec:conclusion}

In this work, we introduced \textit{Turk-LettuceDetect}, a hallucination detection models specifically designed for Turkish RAG applications. Our approach adapts the token-level classification methodology of LettuceDetect to the linguistic characteristics of Turkish by leveraging three novel achitectured base models fine-tuned on the RAGTruth dataset \cite{Niu2024}. Experimental results show that our results achieves strong performance across multiple tasks including question answering, data-to-text writing, and summarization—while maintaining computational efficiency and supporting long-context inputs up to 8,192 tokens \cite{Warner2024}.

Our evaluation reveals critical insights with important implications for multilingual hallucination detection. The \textit{modernbert-base-tr-uncased-stsb-HD} model demonstrates superior performance in summary, qa and whole datasets while \textit{lettucedect-210m-eurobert-tr-v1} model excel in data2txt. Task-dependent behavior patterns show summarization as the most challenging domain, highlighting the need for task-specific detection strategies. Multilingual transfer learning proves highly effective, with EuroBERT-based models achieving robust cross-lingual generalization without full in-language retraining. LLMs (GPT-4.1, Mistral) exhibit high recall but consistently low precision, reinforcing the necessity of dedicated detection mechanisms. Our encoder-based fine-tunel models offers a favorable efficiency-accuracy trade-off with vary range of model sizes such as 135M, 210M and 305M, enabling token-level detection that improves interpretability for real-world applications.

These results establish \textit{Turk-LettuceDetect} models as a foundational step toward robust hallucination detection in Turkish RAG systems. Our work fills a critical gap in multilingual RAG evaluation and supports future research in low-resource, morphologically complex languages. We plan to publicly release our models to encourage reproducibility and broader adoption in real-time, trustworthy RAG pipelines.

\section*{Acknowledgment}
This study is supported by \textbf{GSI Attorney Partnership}. The authors would also like to express their gratitude for the valuable insights and support provided throughout the research process.

\renewcommand{\refname}{References}

\end{document}